\documentclass[sigconf]{acmart}

\copyrightyear{2022}
\acmYear{2022}
\setcopyright{acmcopyright}
\acmConference[CIKM '22] {Proceedings of the 31st ACM International Conference on Information and Knowledge Management}{October 17--21, 2022}{Atlanta, GA, USA.}
\acmBooktitle{Proceedings of the 31st ACM International Conference on Information and Knowledge Management (CIKM '22), October 17--21, 2022, Atlanta, GA, USA}
\acmPrice{15.00}
\acmISBN{978-1-4503-9236-5/22/10}
\acmDOI{10.1145/3511808.3557475}

\usepackage{stfloats}
\usepackage{multirow}
\usepackage[ruled]{algorithm2e}
\usepackage{hyperref}
\hypersetup{hidelinks,
	colorlinks=true,
	allcolors=black,
	pdfstartview=Fit,
	breaklinks=true}

\begin{document}

\title{Towards Federated Learning against Noisy Labels via Local Self-Regularization}


\author{Xuefeng Jiang, Sheng Sun, Yuwei Wang, and Min Liu}
\affiliation{%
  \institution{
    Institute of Computing Technology, Chinese Academy of Sciences, Beijing, China\\
    University of Chinese Academy of Sciences, Beijing, China\\}
  \city{}
  \country{}
}
\email{{jiangxuefeng21b, sunsheng, ywwang, liumin}@ict.ac.cn}



\begin{abstract}
Federated learning (FL) aims to learn joint knowledge from a large scale of decentralized devices with labeled data in a privacy-preserving manner. 
However, since high-quality labeled data require expensive human intelligence and efforts, data with incorrect labels (called noisy labels) are ubiquitous in reality, which inevitably cause performance degradation. 
Although a lot of methods are proposed to directly deal with noisy labels, these methods either require excessive computation overhead or violate the privacy protection principle of FL.
To this end, we focus on this issue in FL with the purpose of alleviating performance degradation yielded by noisy labels meanwhile guaranteeing data privacy. 
Specifically, 
we propose a Local Self-Regularization method, which effectively regularizes the local training process via implicitly hindering the model from memorizing noisy labels and explicitly narrowing the model output discrepancy between original and augmented instances using self distillation.
Experimental results demonstrate that our proposed method can achieve notable resistance against noisy labels in various noise levels on three benchmark datasets. In addition, we integrate our method with existing state-of-the-arts and achieve superior performance on the real-world dataset Clothing1M. The code is available at \href{https://github.com/Sprinter1999/FedLSR}{https://github.com/Sprinter1999/FedLSR}.
\end{abstract}





\maketitle

\section{Introduction}

The pervasiveness of large end devices (e.g. personal mobile phones and IoT devices), has contributed to the drastically increasing scale of data generating from distributed clients among the network. 
Federated learning (FL) is a new distributed learning paradigm that enables a global model to be trained collaboratively by multiple clients while keeping the private data decentralized on devices. 

\begin{figure*}[t]
    \centering
    \includegraphics[width=\textwidth]{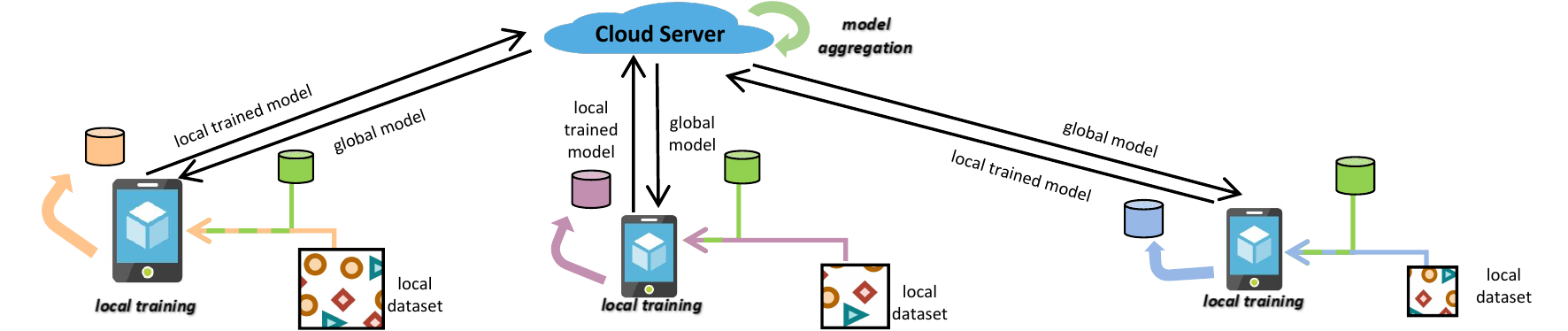}
    \caption{An illustration of the common workflow of federated learning. The global model is updated by aggregating local trained models of a set of selected clients. Note that the local datasets can contain data with noisy labels.}
    \label{fig:flsys}
\end{figure*}

As shown in Figure \ref{fig:flsys}, conventional FL \cite{McMahanMRHA17} process mainly consists of two periods: 1) the selected clients perform local training process on private labeled data after obtaining the distributed global model, and then upload the updated local trained models to the server; 2) the server synchronizes and aggregates these local trained models to obtain the updated global model for next round.
This above process continues until the global model converges or within limited communication rounds. 

Many existing works towards FL have achieved wide success in dealing with four main challenges when deploying a practical FL system, including statistical heterogeneity, systems heterogeneity, communication efficiency, and privacy concerns as referred in \cite{DBLP:journals/spm/LiSTS20}.
But most existing works are based on an important assumption that the raw data owned by clients are perfectly labeled.   
In practice, it is hard to guarantee the label correctness of collected training data, since high-quality annotations requires expensive human intelligence and efforts. 
Moreover, data are very likely to contain incorrect labels (a.k.a noisy labels) in FL, since labels are usually produced independently by clients with various label-generating methods, such as filtering images' surrounding context \cite{mopro} or exploiting machine-generated labels \cite{DBLP:journals/corr/abs-1811-00982}. 

Data with noisy labels inevitably cause the model performance degradation. 
In detail, Arpit et al. propose the memorization effect of deep network \cite{ArpitJBKBKMFCBL17} which indicates that data with correct labels fit before data with incorrect labels, thus the model performance first rises up and then gradually drops during the training process.
Although learning on data with noisy labels \cite{natarajan2013learning} has been widely studied in the data centralized setting, most existing works cannot be straightforwardly applied to FL, due to unbearable computation burden and exorbitant communication overhead for resource-constrained devices. 
For example, \cite{DBLP:conf/cvpr/NishiDRH21,DBLP:conf/icml/JiangZLLF18,DBLP:conf/iclr/LiSH20} perform computation-heavy procedures for noise cleaning \cite{mopro} , which brings non-negligible synchronization cost during conducting the server-side model aggregation, and thus negatively affects global model convergence. 
While other methods cannot satisfy some unique characteristics of FL.  
For example, \cite{DBLP:conf/nips/LiuNRF20} requires direct access to all data samples in the early learning process, but the server only selects a small fraction of clients to conduct local training on these clients' private data in FL. 

To reduce the negative effect caused by noisy labels in FL, most existing researches \cite{chen2020focus,tuor,DBLP:journals/tvt/YangQWZZ22} utilize an auxiliary dataset with perfect labels to identify the noise level of clients or conduct the sample-level selection for training.
However, since it is hard and impractical to obtain a pre-defined, fixed and perfectly-labeled auxiliary dataset, these methods fail to tackle the issue of training on data with noisy labels in reality.
Fortunately, Yang et al. \cite{yang2022robust} firstly propose an auxiliary-dataset-free FL method, which collects local classwise feature centroids from clients to form global classwise feature centroids as global supervision to effectually deal with the noisy labels. 
However, this method can bring about underlying privacy leakage risks, since these centroids transimitted between clients and the server can be inversely utilized to reveal some sensitive information about the private data. 
Therefore, it is necessary to obtain extra reliable supervision for dealing with noisy labels without compromising data privacy.  

Data augmentation is a low-cost and widely-utilized technique to effectively obtain extra supervision information, and has been applied in many machine learning problems such as supervised learning \cite{DBLP:conf/nips/KrizhevskySH12,DBLP:conf/cvpr/XieLHL20}, semi-supervised learning \cite{DBLP:conf/iclr/LaineA17,DBLP:conf/nips/BerthelotCGPOR19,DBLP:conf/nips/XieDHL020}, self-supervised contrastive learning \cite{DBLP:conf/icml/ChenK0H20,DBLP:conf/cvpr/He0WXG20,DBLP:conf/iccv/Banani021} and many other domains. Nevertheless, few works have explored the utilization of data augmentation technique in the scope of learning on data with noisy labels \cite{DBLP:conf/cvpr/NishiDRH21}.
In this work, we focus on the local training process of FL, and utilize data augmentation technique to obtain extra supervision, so as to promote performance against noisy labels without violating the privacy principle. 
Considering the presence of data with noisy labels, our intuition is three-fold: 1) The prediction for the augmented sample should be close to the prediction for the original sample. 2) Model predictions can be more close to the ground truth than the corresponding noisy labels. 3) The proposed approach should be privacy-preserving to be applied in the practical FL system.

Following this intuition, we propose a method named Local Self-Regularization to tackle the issue of training on data with noisy labels in FL, which tries to mitigate the performance degradation caused by noisy labels in the premise of protecting data privacy.
Specifically, we implicitly regularize local training by enhancing model discrimination confidence to prevent it from memorizing noisy labels,
 and further utilize self knowledge distillation technique to explicitly regularize the model output discrepancy between original and augmented instances.
We provide some insights to deal with noisy labels in FL through the following contributions:
\begin{itemize}
  \item We empirically show that in the presence of noisy labels, the memorization effect of deep network proposed in data centralized setting still exists in FL, which brings degradation to the global model performance.
  \item We propose an auxiliary-dataset-free and privacy-preserving FL method named Local Self-Regularization, which implicitly regularizes the model from memorizing noisy labels and explicitly regularizes the model output discrepancy between original and augmented instances. 
  \item We present the effectiveness of our method through extensive experiments on three benchmark datasets compared to the state-of-the-arts in various noise levels.
  In addition, our method has potential to incorporate with other existing methods to further improve performance, and this insight is also verified on the real-world Clothing1M dataset.
\end{itemize}
\section{Related Works}
\label{rw}
\subsection{Federated Learning}
Federated learning \cite{McMahanMRHA17} is a new distributed machine learning paradigm, which has drawn much attention in recent years, 
since data originating from the increasing number of devices gets larger. 
FL aims to fully exploit decentralized data of local devices on basis of data privacy and acceptable communication overhead, and learn a global model via interacting model parameters between clients and the server. 
As \cite{DBLP:journals/spm/LiSTS20} stated, many studies tackle with these underlying issues when deploying practical FL system : 1) statistical heterogeneity challenge caused by non-IID and unbalanced data among clients; 2) systems heterogeneity challenge caused by limited on-device storage, computing capacity and network connectivity; 3) communication efficiency over the FL system; 4) privacy-preserving challenge for increasing security concerns. 
FedProx \cite{DBLP:conf/mlsys/LiSZSTS20}, SCAFFOLD \cite{DBLP:conf/icml/KarimireddyKMRS20}, FedAvgM \cite{DBLP:conf/eccv/HsuQ020} and other works \cite{zhao2018federated,DBLP:conf/iclr/LiHYWZ20} towards the convergence of global model focus on the statistical heterogeneity. 
FedAsync \cite{xie2019asynchronous}, Astraea \cite{DBLP:journals/tpds/DuanLCLTL21}, STC \cite{sattler2019robust} and other works tackle with the local computation or network transmit limitation. 
\cite{DBLP:journals/corr/KonecnyMYRSB16,DBLP:conf/nips/WangSLCPW18} utilize advanced methods like quantization or sparsification to significantly reduce the size of communication information between the clients and server. 
There exist other studies focusing on the newly-emerging privacy concerns within FL from the perspective of both attacks and protection \cite{DBLP:conf/nips/ZhuLH19,DBLP:journals/www/HuangDJQWL20,DBLP:journals/jsait/SoGA21a}.

These studies usually have a strong assumption that each client has a local dataset with correct labels. 
However, in the practical FL system, high-quality annotations or labels cannot be guaranteed, because decentralized data are generating from different clients with various label-generating methods. 
For example, in the real-world setting, the labels can be generated by filtering the website's surrounding context of images \cite{
mopro} or adopting machine-generated technique \cite{DBLP:journals/ijcv/KuznetsovaRAUKP20} on clients. 
Therefore, it is necessary to consider the existence of noisy labels in client-side local datasets and further develop an effectual method against noisy labels.
\subsection{Learning on Noisy Labels}
As mentioned above, training on data with high-quality labels is a strong assumption. 
Even commonly-used datasets like CIFAR-10 \cite{krizhevsky2009learning} and Amazon Reviews \cite{DBLP:conf/sigir/McAuleyTSH15} contain noisy labels \cite{northcutt2021pervasive}. 
Many prior works neglect the existence of noisy labels, and believe the large scale of data can overpower the negative effect caused by noisy labels \cite{DBLP:conf/eccv/MahajanGRHPLBM18}. \cite{DBLP:conf/iclr/ZhangBHRV17} points out deep neural networks have many parameters, and they are capable enough to overfit data with noisy labels, which can cause a performance degradation. 

There are many studies focused on tackling with noisy labels in the data centralized setting.
The most recent advances in learning on noisy labels can be divided into two categories:
(1) selecting or reweighting samples with higher possibility to be correctly-labeled \cite{DBLP:conf/nips/HanYYNXHTS18,DBLP:conf/icml/JiangZLLF18,DBLP:conf/nips/MalachS17}, or (2) obtaining more supervision information from the output of deep networks \cite{DBLP:conf/cvpr/TanakaIYA18,DBLP:conf/cvpr/YiW19,DBLP:conf/iccv/0001MCLY019,DBLP:conf/iclr/LiSH20,DBLP:conf/cvpr/NishiDRH21}. 
For the first category of stategies, many existing methods use peer networks to conduct correct-labeled sample selection or reweighting. 
MentorNet \cite{DBLP:conf/icml/JiangZLLF18} introduces curriculum learning  \cite{bengio2009curriculum} into sample reweighting by pre-training an extra network, which is utilized to set higher weights for correct-labeled inputs during the training process. 
Decoupling \cite{DBLP:conf/nips/MalachS17} maintains two peer networks, and each network updates its parameters using only the samples whose predictions are different in the two networks. 
Following the finding that the sample with lower loss has higher possibility to be correctly labeled \cite{ArpitJBKBKMFCBL17}, Co-teaching  \cite{DBLP:conf/nips/HanYYNXHTS18} also maintains two peer networks, and each network trains the samples with lower loss given by another peer network. 

For another category of strategies, many methods focus on obtaining extra supervision from model output. 
Wang et al. \cite{DBLP:conf/iccv/0001MCLY019} propose that model predictions can reflect the ground truth to a certain degree, and design a robust loss function named Symmetric Cross Entropy. \cite{DBLP:conf/nips/ZhangS18}  applies a Box-Cox transformation to model output and also designs a robust loss term.
\cite{DBLP:conf/cvpr/PatriniRMNQ17} performs loss correction by estimating the overall probability of each class being corrupted into another.  \cite{DBLP:conf/cvpr/TanakaIYA18} points out a relatively bigger learning rate can mitigate the overfitting to noisy labels, and tries to alternately update model parameters and dataset's labels by averaging the deep network's output predictions generated in previous training epochs. 
On this basis, PENCIL \cite{DBLP:conf/cvpr/YiW19} introduces model output logits into loss modeling and updates the estimated label by online gradient descent. 
More recently, DivideMix \cite{DBLP:conf/iclr/LiSH20} combines multiple techniques including Co-teaching, MixUp \cite{mixup}, and MixMatch \cite{DBLP:conf/nips/BerthelotCGPOR19}. 
AugDesc  \cite{DBLP:conf/cvpr/NishiDRH21} uses weak augmentation for loss modeling and stronger augmentation for gradient descent to optimize the training strategy. 

Many above-mentioned works \cite{DBLP:conf/iclr/LiSH20, DBLP:conf/cvpr/NishiDRH21, DBLP:conf/icml/JiangZLLF18} either perform heavy computation procedures or cannot meet some unique characteristics of FL, thus cannot be directly transferred into FL. 
Since clients have limited computation and storage capacity, and only a small fraction of clients are selected to perform local training on private data considering efficiency \cite{McMahanMRHA17}. 
Complex on-device training require excessive training cost and bring non-negligible synchronization time in each round, which can evidently reduce the user interaction experience.
For example, ELR \cite{DBLP:conf/nips/LiuNRF20} requires access to all samples to conduct early-learning loss regularization. 
AugDesc \cite{DBLP:conf/cvpr/NishiDRH21} utilizes reinforcement learning based AutoAugment \cite{DBLP:conf/cvpr/CubukZMVL19} technique. 
Since the overall data distribution cannot be estimated from the server or any single client, AutoAugment technique cannot be directly applied to FL due to the potential performance degradation \cite{DBLP:journals/corr/abs-2111-11066}.

In FL, most existing works focus on the measurement of the possible extent of noisy labels in each client's local dataset to mitigate this issue. FOCUS \cite{chen2020focus} keeps an auxiliary dataset at the server to measure the label noise level 
and reduce the weights of low-quality clients for model aggregation to reduce the negative effect caused by noisy labels. 
Tuor et al. \cite{tuor} use a benchmark model trained on a pre-defined task-specific benchmark dataset, to evaluate the relevance of data at each client and select the data with relatively high relevance. 
Yang et al. \cite{DBLP:journals/tvt/YangQWZZ22} propose to leverage the measured noise ratio of each client based on a common clean validation dataset to guide client selection. 
However, it can be an unrealistic assumption to collect a fixed task-specific auxiliary dataset with perfect labels at the server side in practice, since collecting a high-quality auxiliary dataset requires intricate human intelligence and high cost such as medical diagnoses.
Moreover, from the perspective of continual learning \cite{delange2021continual}, a fixed auxiliary dataset cannot capture the gradual increment of the emerging decentralized data. 
Besides, these methods fail to fully utilize clients' datasets as they ignore the underlying knowledge of data with noisy labels. 
More recently, to our best knowledge, \cite{yang2022robust} is the first work to directly deal with noisy labels, and tries to obtain global supervision by collecting local classwise feature centroids to form global classwise feature centroids as global supervision. These centroids are the model intermediate output, which are regarded as important and sensitive information in FL \cite{wei2022vertical} and can be inversely utilized by malicious attackers to reveal some private information of the raw data.
Thus, the classwise features directly transmitted between clients and the server can cause the issue of privacy leakage.
\subsection{Knowledge Distillation}
Knowledge distillation \cite{DBLP:journals/corr/HintonVD15,cristian2006model} aims to transfer knowledge from a complex teacher model to a lightweight student model by minimizing the output or intermediate features of these two networks. Although knowledge distillation has successfully applied in many fields, such as model heterogeneity in FL \cite{DBLP:journals/corr/abs-1910-03581} and domain adaption in transfer learning \cite{DBLP:conf/wacv/Nguyen-MeidineB21}, obtaining a complex teacher model still requires a large amount of extra training overhead.
To further improve efficiency for knowledge transferring, self knowledge distillation \cite{wang2021knowledge,DBLP:conf/nips/MobahiFB20} (or called self distillation) aims to utilize knowledge from model itself without the involvement of an explicit teacher network.
Existing methods usually exploit knowledge from model itself \cite{DBLP:conf/ranlp/HahnC19} or distill knowledge between different network layers \cite{DBLP:conf/iccv/ZhangSGCBM19,DBLP:conf/iccv/HouMLL19}. 
Different from these above-mentioned methods, we utilize self distillation at the instance level to minimize the discrepancy of the model output between the original and augmented samples.
\section{Method}
In this section, we first provide the problem definition, briefly introduce the memorization effect of deep networks and recent methods which obtain extra supervision from the model output.
We then conduct a preliminary experiment to show that the memorization effect of deep networks still exists in FL, and data augmentation can mitigate the negative effect caused by noisy labels to a limited degree. After that, we elaborate our method, which intends to obtain extra supervision from the model output and augmented instance.

\subsection{Problem Definition}
\label{warmup}
\label{Met}
\begin{figure*}[t]
    \centering
    \includegraphics[width=\textwidth]{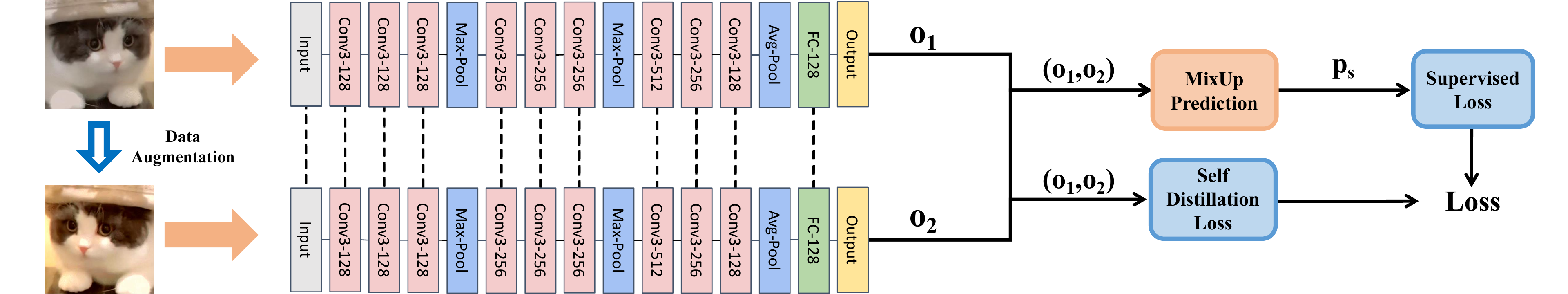}
    \caption{A brief illustration of our method Local Self-Regularization (LSR). We feed the original and augmented instance into the same network and produce output logits $o_1$ and $o_2$. They are utilized in two components. MixUp prediction is to implicitly regularize training by enhancing model discrimination confidence to avoid overfitting noisy labels, and Self distillation is to explicitly regularize the output discrepancy of these two instances.}
    \label{fig:methodd}
\end{figure*}
Without loss of generality, assume the FL system consists of $N$ clients and a server.
Let $\mathcal{S}$ represent the set for all $N$ clients.
Each client $k$ maintains a local dataset with $n_k$ samples denoted as $\mathcal{D}_{k}=\left\{\left(x_{k}^{i}, y_{k}^{i}\right)\right\}_{i=1}^{n_{k}}$, and $\mathcal{D}=\left\{\mathcal{D}_{k}\right\}_{k=1}^{N}$ with $M$-class, some of which have noisy labels.
In each communication round $t$, a subset of clients $\mathcal{S}_t$ are selected to receive the global model $w_t$ from the server, and perform local training for $E$ epochs. 
In FedAvg \cite{McMahanMRHA17}, the selected client $k$ loads the global model $w_t$ to its local model $w_t^k$ and trains on $\mathcal{D}_k$ to reduce cross entropy (CE) loss, which can be given by:
\begin{equation}
\label{CE}
Loss_{CE} = CE(f_{\theta}(x),y),
\end{equation}
where $w_t^k$ is denoted as $\theta$ and $f_{\theta}$ is the function approximated by the deep network. The model output logits for sample $x$ is $f_{\theta}(x)$, and the given label is $y$ (which can be a noisy label).
After finishing local training, each client $k$ sends the updated local model $w_t^k$ back to the server. 
Then the server aggregates all the uploaded local models to obtain a new global model $w_{t+1}$ for the next global round. 


As mentioned above, the memorization effect of deep networks \cite{ArpitJBKBKMFCBL17}  indicates that the deep network model tends to fit the correctly-labeled data before noisy-labeled data during training, and the samples with lower loss are more likely to be correctly-labeled. 
So there exists a short warm-up period $t_w$ during which the model performance increases at early learning process. Then the model develops a basic discrimination ability.
But by directly using the given labels (which can be wrong) as the supervision signal, the model tends to approximate the labels to reduce the CE loss. 
As a result, the deep network gradually memorizes noisy labels, thus resulting in model performance degradation. 
To avoid overfitting noisy labels, many existing works exploit this effect by properly adding more supervision from the model output.

Based on this, a sample selection method Co-teaching \cite{DBLP:conf/nips/HanYYNXHTS18}, maintains two networks $f$ and $g$ with different parameters, selects and feeds samples with lower CE loss to the other peer network, and mutually trains each network on the data which is deemed as correctly-labeled by the other network.
In detail, for each batch of input data $\overline{\mathcal{D}}$, 
Co-teaching selects the correctly-labeled group $\mathcal{D}^{\prime}$, which is regarded as correctly-labeled by the peer network. 
The selection method of $\mathcal{D}^{\prime}$ for each peer network can be formulated as below:
\begin{equation}
\overline{\mathcal{D}}_{f}=argmin_{\mathcal{D}^{\prime}:\left|\mathcal{D}^{\prime}\right| \geq R(T)|\overline{\mathcal{D}}|} \ell\left(f, \mathcal{D}^{\prime}\right);
\end{equation}
\begin{equation}
\overline{\mathcal{D}}_{g}=argmin _{\mathcal{D}^{\prime}:\left|\mathcal{D}^{\prime}\right| \geq R(T)|\overline{\mathcal{D}}|} \ell\left(g, \mathcal{D}^{\prime}\right),
\end{equation}
where $R(T)$ is a threshold for the ratio of samples to place into $\mathcal{D}^{\prime}$ empirically determined by the estimation of noise ratio for the entire dataset. 
Network $f$ selects more reliable samples $\overline{\mathcal{D}}_{f}$ with lower CE loss for its peer network $g$, aiming to use more reliable supervision information for guiding the peer network $g$ update. And network $g$ does the same for network $f$.

Besides, Wang et al. \cite{DBLP:conf/iccv/0001MCLY019} design a robust loss function by emphasizing model outputs to deal with noisy labels.
Symmetric Cross Entropy (Symmetric CE) is developed to obtain more reliable supervision from model itself, which can be formulated as:
\begin{equation}
Loss_{\text{Symmetric CE}}=\alpha*CE(f_{\theta}(x),y)+\beta*CE(y,f_{\theta}(x)),
\end{equation}
where $\alpha$ and $\beta$ are trade-off coefficients for loss modeling.
There are two terms incorporated in Symmetric CE, $CE(f_{\theta}(x),y)$ is CE, and $CE(y,f_{\theta}(x))$ is the reverse of CE (RCE). 
As stated in  \cite{DBLP:conf/iccv/0001MCLY019}, CE is useful for achieving good convergence, and RCE is noise-tolerant. 
Symmetric CE emphasizes that only using the unreliable labels as supervision to conduct CE can be deficient, and then utilize the model output logits to form RCE as an extra supervision.  

Co-teaching and Symmetric CE are proposed in data centralized setting but can be utilized in local training process of FL. 
In the scope of FL, to deal with noisy labels, Yang et al. \cite{yang2022robust} devote to obtaining global supervision by collecting clients' classwise feature centroids to gradually form global classwise feature centroids and conducting entropy regularization to enhance model predictions. 
The loss function is formulated as below:
\begin{equation}
Loss_{\text {Robust FL}}^{k}=L_{c}^{k}+{\lambda_{cen}}*L_{c e n}^{k}+{\lambda_{e}}*L_{e}^{k},
\end{equation}
where $L_{c}$ denotes CE loss, $L_{cen}$ denotes the difference between global classwise features and local classwise features on client $k$, and $L{e}$ denotes the entropy regularization loss which enforces the model to produce predictions with lower entropy, as \cite{yang2022robust} stated. $\lambda_{cen}$ and $\lambda_{e}$ are trade-off coefficients. 

The above mentioned methods focus on emphasizing model outputs encourages us to explore intuition for our method.
In this work, we follow up the insight of enhancing model output prediction and implicitly regularize the training process to hinder the model from overfitting the noisy labels.  

\subsection{Noisy Labels in Federated Learning}
We conduct a preliminary experiment with FedAvg \cite{McMahanMRHA17} to observe the memorization effect of deep network in FL. 
The basic experimental setting is depicted in section \ref{setup}. We conduct experiments with various noise levels and we show the test accuracy for several kinds of general augmentation methods with no noise or with a neutral ($\epsilon=0.4$) symmetric noise (discussed in \ref{setup}) in here. 

\begin{figure}[htb]
    \centering
    \includegraphics[width=\linewidth]{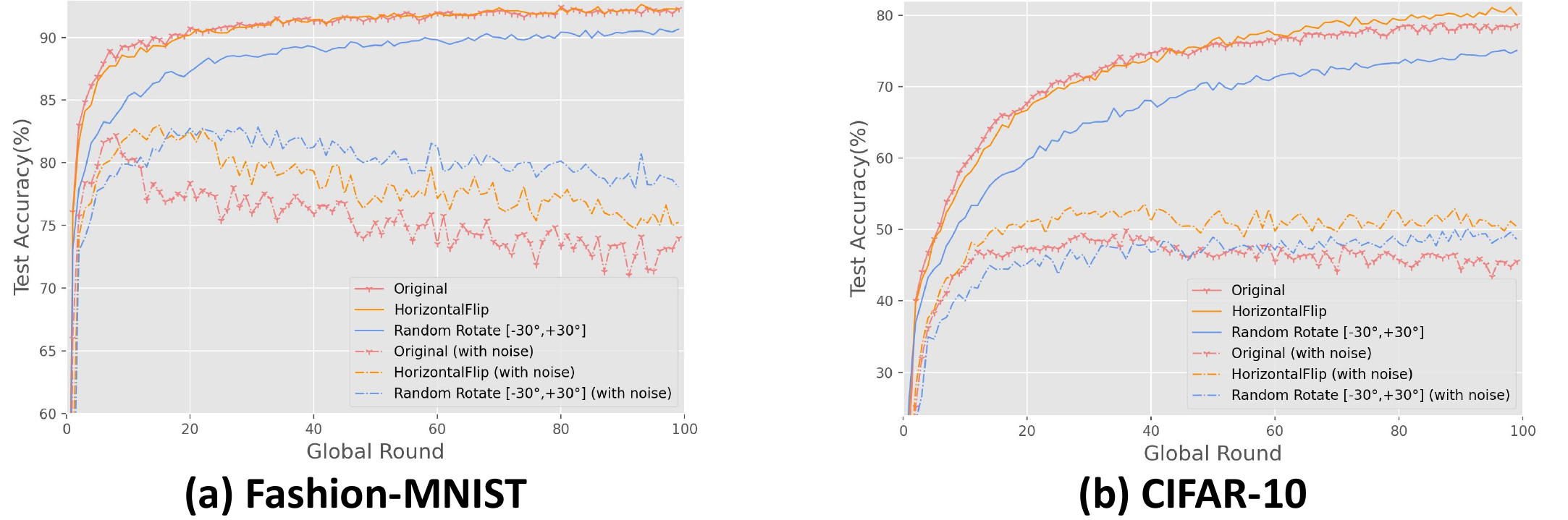}
    \caption{Test accuracy on both original and augmented Fashion-MNIST and CIFAR-10 with symmetric noise ($\epsilon = 0.4$).}
    \label{fig:aug}
\end{figure}

Intuitively, as shown in Table \ref{aug} and Figure \ref{fig:aug}, we can observe that the global model performance first rises up and then gradually drops on the original data domain with noise, and Meanwhile, this phenomenon can still be observed in many other noise settings.
This indicates that the memorization effect of deep network still exists in FL due to the negative knowledge learned from distributed data with noisy labels. 
In the presence of data with noisy labels, data augmentation can improve the model robustness against the noisy labels to a limited degree. 
In Figure \ref{fig:aug}(a), with no noisy labels, training on original domain converges fastest, and training on augmented domain with random rotation within 30 degrees converges slowest. 
While training on data with 40\% symmetric noise, applying random rotation within 30 degrees shows better robustness against noisy labels. 
The warm-up period $t_w$ (discussed in section \ref{warmup}) is about 20 rounds on Fashion-MNIST (considering both original and augmented domains). 
This encourages us to further utilize data augmentation for improving robustness against the noisy labels.
\begin{table}[htb]
\caption{Test accuracy (\%) on benchmark datasets with various augmentation methods against symmetric noisy labels. }
\label{aug}
\label{tab:aug}
\begin{tabular}{c|cccc}
\cline{1-5}
Dataset                       & \multicolumn{2}{c|}{Fashion-MNIST} & \multicolumn{2}{c}{CIFAR-10} \\ \hline
Adding Noise                  & False & \multicolumn{1}{c|}{True}  & False         & True         \\ \hline
Original                      & 92.13 & \multicolumn{1}{c|}{72.75} & 78.58         & 45.72        \\ \hline
Random Horizontal Flip      & 92.26 & \multicolumn{1}{c|}{75.40} & 80.31         & 49.14        \\ \hline
Random Rotate  {[}-5°,+5°{]}   & 91.77 & \multicolumn{1}{c|}{74.69} & 77.77         & 47.58        \\ \hline
Random Rotate {[}-15°,+15°{]} & 91.21 & \multicolumn{1}{c|}{77.13} & 76.68         & 48.14        \\ \hline
Random Rotate {[}-30°,+30°{]} & 90.46 & \multicolumn{1}{c|}{78.74} & 74.64         & 49.02        \\ \hline
\end{tabular}
\end{table}
\subsection{Local Self-Regularization Method}
\label{actualmethod}

Our method mainly focuses on optimizing client-side local training process by implicitly regularizing model predictions to hinder the trained model from overfitting noisy labels, and explicitly regularizing output discrepancy of original and augmented instances.

\textbf{MixUp prediction}
When the selected client $k$ receives the global model $w_t$ at the $t$-th round, $k$ loads this model as its local model $w_t^k$. 
By conducting data augmentation, we obtain two versions of each sample, the original $x$ and the augmented $\hat{x}$. 
They are fed into the deep network to yield two output logits $o_1$ and $o_2$ , and the corresponding predictions (or soft target) $p_1$ and $p_2$ by conducting SoftMax on $o_1$ and $o_2$. 
To simultaneously use these two predictions, we randomly sample $\lambda$ from Beta distribution to mix $p_1$ and $p_2$ up, and get the mixed prediction $p$ :
\begin{equation}
\label{mixed}
        p = \lambda*p_1 + (1-\lambda)*p_2.
\end{equation}
\begin{figure}[b]
    \centering
    \includegraphics[width=\linewidth]{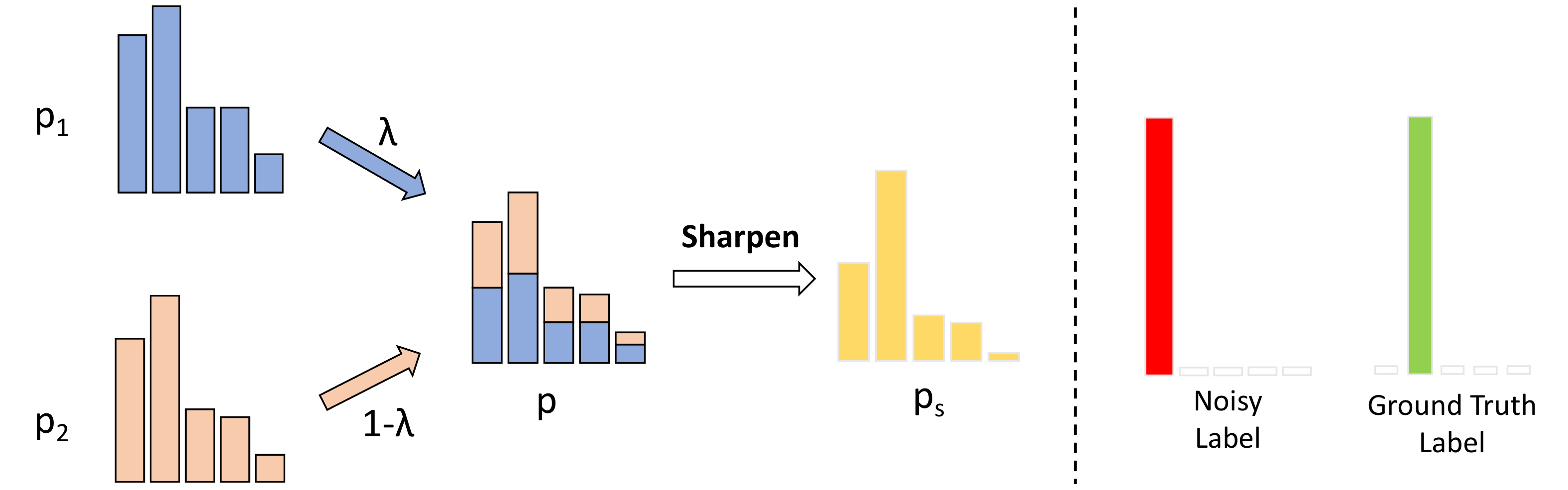}
    \caption{An intuitive understanding for MixUp prediction. In this scenario, the model gives correct prediction but the given label $y$ is wrong (namely $y$ is a noisy label). By conducting sharpening operation, $CE(p_{s},y)$ is larger than $CE(p_1,y)$, $CE(p_2,y)$ and $CE(p,y)$, which implicitly adds the difficulty for overfitting the noisy label $y$.}
    \label{fig:sharpen}
\end{figure}
The smooth assumption \cite{chapelle2009semi} in many semi-supervised learning methods indicates that the classifier’s decision boundary should not pass through high-density regions of the marginal data distribution. 
Entropy minimization \cite{DBLP:conf/nips/GrandvaletB04} can help to meet this assumption via adding a loss term for minimizing the entropy of model predictions. 
Enlightened by this assumption, but different from previous works, to enhance the model prediction confidence, we directly apply a sharpening operation on the mixed prediction $p$ to obtain the sharpened prediction $p_{s}$:
\begin{equation}
p_{s,i}=\operatorname{Sharpen}(p, T)_{i}:=p_{i}^{\frac{1}{T}} / \sum_{j=1}^{M} p_{j}^{\frac{1}{T}},
\end{equation}
where $i$ denotes the $i$-th class, and $T$ is the sharpening temperature. 
Then We use $p_{s}$ instead of logits in Eq. \ref{CE} to compute classification loss, which can be expressed as:
\label{cls}
   $ Loss_{cls}=CE(p_{s},y).$

Note that the sharpening operation is also used in Mixmatch \cite{DBLP:conf/nips/BerthelotCGPOR19} to produce pseudo labels of unlabeled data, and to further conduct MixUp \cite{mixup} on the overall data for semi-supervised learning. 
Here, we directly use this sharpened prediction $p_{s}$ to conduct loss modeling (cross entropy loss). 
The underlying motivation is to enforce the model to make predictions with more confidence (lower entropy) on each sample. 
The original model output logits tend to approximate the given label (which can be noisy) to reduce the CE loss, thus the network forms some wrong patterns for discrimination. 
But the sharpened prediction can enhance the model's prediction confidence, which hinders the deep network from memorizing noisy labels via adding difficulty to reduce the CE loss.
This acts as an implicit regularizing method to avoid overfitting noisy labels.
\begin{algorithm}[b] 
    \caption{Local training with Local Self-Regularization}
    \LinesNumbered
    \KwIn{client $k$ , global model $w_t$ , coefficient $\gamma$ }
    \KwOut{local trained model $w_t^k$}
    $w_t^k \gets w_t$\; 
    \For{each local epoch $i$ from 1 to E}{
        \For{each batch $(x,y)$ }{
        $o_1, o_2 = f(x;w_t^k), f(Augment(x);w_t^k)$\;
        $p_1, p_2 = SoftMax(o_1), SoftMax(o_2)$ \; 
        $\lambda \sim Beta(1,1)$ \;
        $p = \lambda*p_1 + (1-\lambda)*p_2$  \;
        $p_{s} = Sharpen(p,T)$ \;
        $Loss_{cls}=CrossEntropy(p_{s},y)$ \;
        $Loss_{reg} = SelfDistillation(o_1,o_2, T_d)$ \;
        $Loss = Loss_{cls} + \gamma * Loss_{reg} $ \;
        Update $w_t^k$ with $Loss$ \;
        }
    }

\end{algorithm}

\textbf{Self distillation}
Intuitively, each pair of the original sample $x$ and augmented sample $\hat{x}$ belongs to the same class, which can be utilized to obtain a self supervision information. 
In previous studies towards knowledge distillation, L1 loss (L1 distance), L2 loss (L2 distance), cosine similarity, JensenShannon (JS) divergence \cite{DBLP:conf/cvpr/YinMALMHJK20} and other metrics can be used to measure the discrepancy or similarity between the teacher and student model outputs (e.g. last layer's output logits or intermediate features) to form the distillation loss. 
In our method, the teacher and the student model are the same network, and we use the two output logits of $x$ and $\hat{x}$ ($o_1$ and $o2$) to conduct instance-level self distillation. 
We choose either JensenShannon (JS) divergence or L1 distance here to represent the discrepancy between the output $o_1$ and $o_2$ of original data $x$ and augmented $\hat{x}$. We first use a SoftMax function with distillation temperature $T_d$ to modify $o_1$ and $o_2$, and obtain the corresponding $q_1$ and $q_2$ as follows:
\begin{equation}
q_{1,i}, q_{2,i}=\frac{\exp \left(o_{1,i} / T_d\right)}{\sum_{j} \exp \left(o_{1,j} / T_d\right)}, \frac{\exp \left(o_{2,i} / T_d\right)}{\sum_{j} \exp \left(o_{2,j} / T_d\right)};
\end{equation}
where both $i$ and $j$ represents the output logits for $i-th$ and $j-th$ class (M classes in total). 
For JS divergence, the self distillation loss term can be formulated as:
\begin{equation}
\label{JSdiv}
Loss_{reg} = J S(q_1, q_2)=\frac{1}{2}(K L(q_1 \| U)+K L(q_2 \| U)),
\end{equation}
where KL means Kullback-Leibler divergence and $U=\frac{1}{2}({q_1}+{q_2})$ . 
This acts as an explicit regularization to obtain instance-level supervision. 
Then we form the final client-side loss function:
\begin{equation}
    Loss = Loss_{cls} + \gamma * Loss_{reg},
\end{equation}
where $\gamma$ is the trade-off coefficient received from the server in each round. 
Note that we linearly increase this coefficient from 0 to $\gamma$ in the warm-up rounds $t_w$ (discussed in \ref{warmup}), so as to prioritize fitting task in the early learning process and then gradually optimize the self distillation loss.

\textbf{Model aggregation}
\label{modelagg}
We follow FedAvg \cite{McMahanMRHA17} to execute model aggregation at the server side. 
The server aggregates the uploaded local model parameters as follows:
    $w_{t+1}=\sum_{k \in S_t} \frac{n_{i}}{n} w_{t}^k,$
where $w_{t+1}$ denotes the global model parameter for the next $t+1$-th round, $w_{t}^k$ denotes the local model trained on client $k$ at the $t$-th round.
$n_i$ and $n$ indicate the number of local data of selected client $i$ and the total number of data of all selected clients, respectively.

\section{Experiments}
\label{exp}
\subsection{Experimental Setup}
\label{setup}
The basic experimental setting is as follows. We set the number of total clients $N$ to 100. Each client's local dataset has the same number of samples. We select local iteration epoch $E$ of each client to 5, and local batchsize to 60. 

\begin{figure}[htb]
    \centering
    \includegraphics[width=\linewidth]{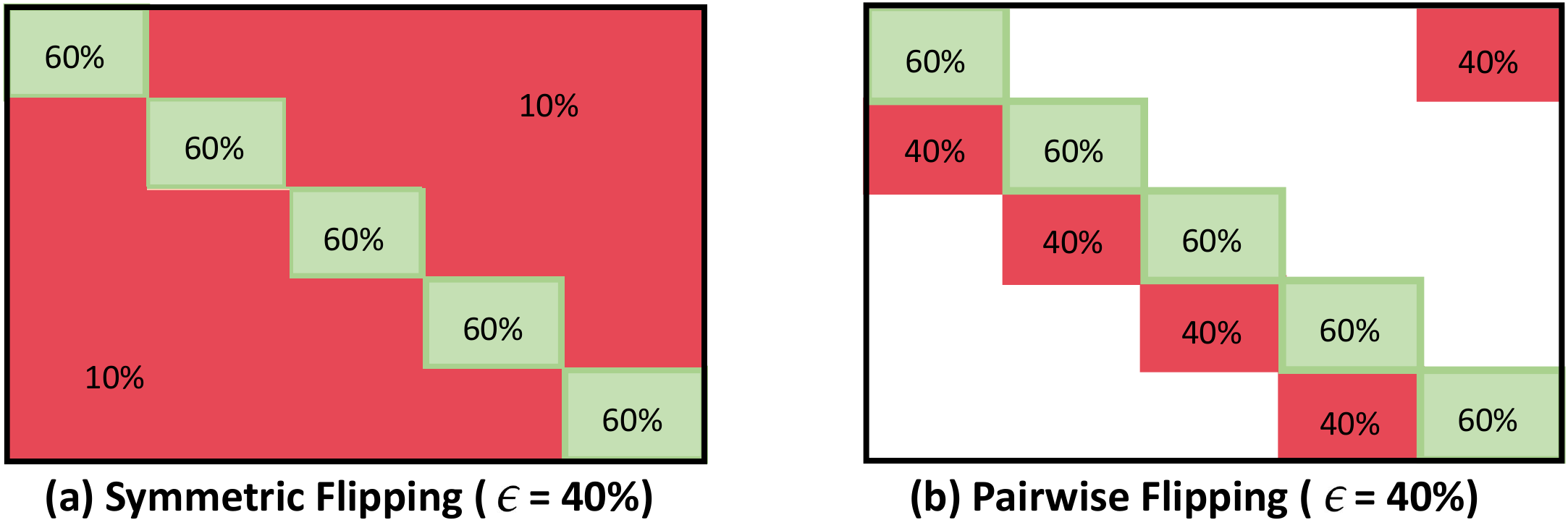}
    \caption{Transition matrices of different noise types (using 5 classes as an example). The green and red grids represent the percentage of samples that are correctly labeled to the ground truth class, and the percentage of samples that are incorrectly labeled to other classes, respectively.}
    \label{fig:case}
\end{figure}

\textbf{Datasets with noisy labels} 
We firstly perform extensive experiments on MNIST \cite{DBLP:journals/pieee/LeCunBBH98}, Fashion-MNIST \cite{DBLP:journals/corr/abs-1708-07747} and CIFAR-10 \cite{krizhevsky2009learning}, three benchmark image datasets. 
These datasets contain $M=10$ categories of images for classification. MNIST and Fashion-MNIST has 60K images for training and 10K images for testing of size 28$\times$28. 
CIFAR-10 has 50K images for training and 10K images for testing of size 32$\times$32. 
We generate synthetic noisy labels by replacing the original labels in these datasets with two typical types of noise: symmetric flipping \cite{DBLP:conf/nips/RooyenMW15} and pairwise flipping \cite{DBLP:conf/nips/HanYYNXHTS18}. 
We select various levels of the noise ratio $\epsilon$. 
Note that when the noise ratio is fixed, pair flipping case is usually harder than the symmetric flipping case. 
In Figure \ref{fig:case}(b), the true class only has 20\% more correct-labeled instances over wrong ones for each class. 
However, the true class has 50\% more correct instances in Figure \ref{fig:case}(a).

In detail, we first divide the dataset into 10 parts according to the original classes, and then evenly inject equivalent amount of noisy labels to samples for each class according to the designate noise type and ratio, with fixed random seeds for a fair comparison. 
Next, we distribute the samples with noisy labels to clients according to the client's ground truth label distribution. 
For example, for IID data partitioning, the ground truth class distribution of each client's dataset nearly follows uniform distribution, but the actual class distribution of each client's dataset is transitioned according to the corresponding noise type and noise ratio. 
For the non-IID setting, each client actually has 5 random classes of samples in reality, but the labels are partially transitioned to noisy labels. 

We further perform experiments on a real-world dataset Clothing1M \cite{DBLP:conf/cvpr/XiaoXYHW15}. 
It is a large-scale dataset with $M=14$ categories, which contains 1 million images of clothing with unstructured noisy labels, since it is obtained from several online shopping websites. It is reported that the overall noise level is approximately 61.54\% \cite{DBLP:conf/cvpr/XiaoXYHW15,DBLP:conf/cvpr/NishiDRH21}. 
We randomly divide 1 million images into 100 partitions with the equivalent number of samples, and set them as local clients' datasets. 
Note that in this partitioning, each client's dataset naturally follows the non-IID fashion with unstructured noise, and we use the dataset with correct labels for testing.

\textbf{Implementation details} 
All experiments are implemented by Pytorch \cite{paszke2017automatic} on Nvidia GeForce RTX 2080Ti GPUs. 
For MNIST, Fashion-MNIST and CIFAR-10, we adopt a 9-Layer CNN applied in \cite{DBLP:conf/nips/HanYYNXHTS18,yang2022robust}.  The learning rate is mildly set to 0.15. We select 5 clients in each round (100 rounds in total).
For the Clothing1M dataset, we exploit ResNet-50 \cite{DBLP:conf/cvpr/HeZRS16} pre-trained on ImageNet \cite{deng2009imagenet} by following \cite{DBLP:conf/iclr/LiSH20}. We adopt SGD optimizer with momentum $0.9$, and weight decay $10^{-4}$ for a fair comparison. Unless otherwise specified, we report the average accuracy of the global model over the last 10 rounds with IID partioning.
More detailed experimental setting for Clothing1M is discussed in section \ref{realworldtest}.

\textbf{Baselines} 
We compare with the following baselines: Cross Entropy adopted in FedAvg \cite{McMahanMRHA17}, Co-teaching \cite{DBLP:conf/nips/HanYYNXHTS18}, Symmetric CE \cite{DBLP:conf/iccv/0001MCLY019}, Robust Federated Learning (Robust FL) \cite{yang2022robust} which are detailedly discussed in section \ref{warmup}. 
We basically follow the consistent hyperparameter setting of each baseline according to related papers. 
For Symmetric CE evaluated on Fashion-MNIST, CIFAR-10 and Clothing1M, $\alpha$ and $\beta$ are fixed to 0.1 and 1 respectively. 
For Robust FL on Fashion-MNIST, the hyperparameter setting is the same as that for evaluations on MNIST. 
Note that for Symmetric CE and Co-teaching, we aggregate the weights of local trained models at the server using FedAvg (depicted in section \ref{modelagg}).

\textbf{Hyperparameters}
In our method, the distillation temperature $T_d$ is empirically set to $\frac{1}{3}$ to mitigate negative knowledge from noisy labels, and the sharpening temperature $T$ is set to $\frac{1}{2}$ by directly following MixMatch \cite{DBLP:conf/nips/BerthelotCGPOR19}. $q_1$ and $q_2$ in Eq.\ref{JSdiv} are clamped to [$10^{-6}$,1] before computing self distillation loss for stability.
The warm-up period rounds $t_w$ is set to 10, 20, 40, 10 for MNIST, Fashion-MNIST, CIFAR-10 and Clothing1M. 
We empirically increase the corresponding coefficient $\gamma$ as referred in Table \ref{tab:betaa} as the noise level $\epsilon$ increases, while the performance is not very sensitive to $\gamma$ in most cases (further discussed in section \ref{ablation}). 
As for data augmentation, we apply random rotation within 30 degrees for MNIST and Fashion-MNIST. 
For CIFAR-10 and Clothing1M, we adopt random horizontal flipping and random color distortion used in SimCLR
\cite{DBLP:conf/icml/ChenK0H20}.

\begin{table}[htb]
\caption{Coefficient $\gamma$ selection.}
\label{tab:betaa}
\resizebox{\linewidth}{!}{
\begin{tabular}{|c|ccccc|ccc|}
\hline
Noise Type    & \multicolumn{5}{c|}{Symmetric } & \multicolumn{3}{c|}{Pairwise } \\ \hline
Noise Ratio    & 0.3    & 0.4    & 0.5   & 0.6   & 0.7   & 0.2         & 0.3         & 0.4        \\ \hline
MNIST & 0.15   & 0.20   & 0.25  & 0.30  & 0.80  & 0.40        & 0.60         & 1.00       \\ \hline
Fashion-MNIST & 0.15   & 0.20   & 0.25  & 0.30  & 0.60  & 0.40        & 0.60         & 1.00       \\ \hline
CIFAR-10      & 0.10   & 0.20   & 0.25  & 0.30  & 0.60  & 0.30        & 0.60        & 0.80       \\ \hline
\end{tabular}}
\end{table}

\subsection{Results and Analysis}
\label{results}
\begin{table*}[h]
\centering
\caption{Test accuracy on benchmark datasets with various noise levels. LSR is the proposed Local Self-Regularization method.}
\label{tab:main}
\begin{tabular}{c|c|ccccccccl}
\hline
                 & \textbf{Method}           & \multicolumn{9}{c}{\textbf{Test Accuracy (\%)}}                                                                                                                                                         \\ \cline{2-11} 
\textbf{Dataset} & Noise Type                & \multicolumn{5}{c|}{Symmetric }                                                                & \multicolumn{3}{c|}{Pairwise }                               & \multirow{2}{*}{Avg.} \\ \cline{2-10}
                 & Noise Ratio               & 0.30           & 0.40           & 0.50           & 0.60           & \multicolumn{1}{c|}{0.70}           & 0.20           & 0.30           & \multicolumn{1}{c|}{0.40}           &                       \\ \hline
                 & FedAvg   \cite{McMahanMRHA17}                 & 91.34          & 83.53          & 73.60          & 57.86          & \multicolumn{1}{c|}{42.12}          & 94.27          & 85.52          & \multicolumn{1}{c|}{70.35}          & 74.82                 \\
                 & Symmetric CE    \cite{DBLP:conf/iccv/0001MCLY019}          & 99.10          & 98.91          & 98.54          & 97.77          & \multicolumn{1}{c|}{95.10}          & 99.10          & 98.63          & \multicolumn{1}{c|}{94.13}          & 97.66                 \\
MNIST            & Co-teaching \cite{DBLP:conf/nips/HanYYNXHTS18}               & 98.80          & 98.11          & 97.38          & 95.94          & \multicolumn{1}{c|}{93.84}          & 98.83          & 97.97          & \multicolumn{1}{c|}{94.43}          & 96.91                 \\
                 & Robust Federated Learning \cite{yang2022robust} & 99.07          & 98.92          & 98.84          & 98.44          & \multicolumn{1}{c|}{\textbf{98.40}} & 99.08          & 99.01         & \multicolumn{1}{c|}{\textbf{98.98}} & 98.84                 \\
                 & FedAvg + LSR & \textbf{99.23} & \textbf{99.14} & \textbf{98.91} & \textbf{98.63} & \multicolumn{1}{c|}{98.01}          & \textbf{99.36} & \textbf{99.22} & \multicolumn{1}{c|}{98.49}          & \textbf{98.87}        \\ \hline
                 & FedAvg  \cite{McMahanMRHA17}                  & 80.02          & 72.75          & 62.29          & 49.22          & \multicolumn{1}{c|}{35.87}          & 86.44          & 77.38          & \multicolumn{1}{c|}{63.05}          & 65.88                 \\
                 & Symmetric CE    \cite{DBLP:conf/iccv/0001MCLY019}          & 88.86          & 85.96          & 80.32          & 69.87          & \multicolumn{1}{c|}{49.34}          & 89.99          & 84.51          & \multicolumn{1}{c|}{68.44}          & 77.16                 \\
Fashion-MNIST    & Co-teaching  \cite{DBLP:conf/nips/HanYYNXHTS18}             & 89.22          & 88.11          & 86.82          & 84.43          & \multicolumn{1}{c|}{81.03}          & 90.37          & 87.77          & \multicolumn{1}{c|}{83.03}          & 86.35                 \\
                 & Robust Federated Learning  \cite{yang2022robust}& 88.26          & 87.41          & 85.54          & 84.04          & \multicolumn{1}{c|}{79.22}          & 89.67          & 89.12          & \multicolumn{1}{c|}{88.17}          & 86.43                 \\
                 & FedAvg + LSR & \textbf{90.42} & \textbf{89.73} & \textbf{88.67} & \textbf{87.00} & \multicolumn{1}{c|}{\textbf{82.72}} & \textbf{90.84} & \textbf{90.27} & \multicolumn{1}{c|}{\textbf{88.34}} & \textbf{88.50}        \\ \hline
                 & FedAvg       \cite{McMahanMRHA17}             & 53.78          & 46.06          & 36.93          & 28.45          & \multicolumn{1}{c|}{19.80}          & 66.93          & 58.47          & \multicolumn{1}{c|}{48.04}          & 44.81                 \\
                 & Symmetric CE   \cite{DBLP:conf/iccv/0001MCLY019}           & 64.80          & 56.40          & 47.45          & 34.11          & \multicolumn{1}{c|}{23.97}          & 67.56          & 59.48          & \multicolumn{1}{c|}{45.91}          & 49.96                 \\
CIFAR-10         & Co-teaching    \cite{DBLP:conf/nips/HanYYNXHTS18}           & 70.23          & 66.84          & 62.54          & \textbf{56.25} & \multicolumn{1}{c|}{\textbf{45.28}} & 71.44          & 66.41          & \multicolumn{1}{c|}{57.21}          & 62.03                 \\
                 & Robust Federated Learning \cite{yang2022robust} & 66.29          & 60.38          & 54.05          & 43.18          & \multicolumn{1}{c|}{32.38}          & 69.01          & 61.18          & \multicolumn{1}{c|}{49.71}          & 54.52                 \\
                 & FedAvg + LSR & \textbf{72.10} & \textbf{68.53} & \textbf{64.27} & 55.10          & \multicolumn{1}{c|}{40.61}          & \textbf{73.79} & \textbf{70.66} & \multicolumn{1}{c|}{\textbf{59.40}} & \textbf{63.06}        \\ \hline
\end{tabular}
\end{table*}

The main results are shown in Table \ref{tab:main}. For Co-teaching, we report the average test accuracy of two peer networks. 
Experimental results indicate that our method enables notable resistance against the noisy labels in various noise levels. 
Compared with FedAvg which neglects the existence of noisy labels, other methods usually achieve superior performance. 
For MNIST, our approach is slightly worse than Robust FL with extreme noisy labels. 
It presents the efficacy of properly using global classwise features (namely representations) as an extra supervision to guide training. 
For CIFAR-10, our approach is slightly worse than Co-teaching with extreme symmetric noisy labels. 
It indicates the efficacy of reliable sample selection in the presence of noisy labels especially in extreme cases. 
However, Co-teaching needs a sensitive hyperparameter $R(T)$ (discussed in section \ref{warmup}) to represent the noise level of local datasets, which is hard to priorly estimate in the practical FL system.
While Robust FL collects and transmits classwise feature centroids from client-side datasets to form the overall classwise feature as the global supervision to guide local training process, which can cause privacy leakage because these centroids can be used to inversely reveal the private information of the raw data.
In general, our method utilizes extra reliable supervision and focus on optimizing the local training process without transmitting extra sensitive information in order to further protect data privacy.   

\subsection{Ablation Study}
\label{ablation}

There are two main components in our method. 
MixUp prediction is to implicitly regularize training by enhancing model discrimination confidence to avoid overfitting noisy labels, 
and self distillation is to explicitly regularize the model output consistency at the instance level. 
We conduct the ablation study by removing each component to evaluate the performance degradation on Fashion-MNIST dataset with IID partitioning.

\textbf{MixUp prediction}
We evaluate the effectiveness of MixUp prediction by removing it from our method.
The results are shown in Table \ref{tab:mixup}. 
Our method without MixUp prediction uses the augmented data to expand the original dataset, and uses vanilla CE loss for training. 
The corresponding results indicate that MixUp prediction is a necessary component to guarantee the superior performance of our method. Meanwhile, compared with only sharpening the original prediction $p_1$, randomly sampling $\lambda$ to get the mixed $p$ in Eq.\ref{mixed} and then sharpening $p$ improves robustness in extreme cases.

\begin{table}[htb]
\caption{Test accuracy (\%) for the effect of MixUp Prediction.}
\label{tab:mixup}
\resizebox{\linewidth}{!}{
\begin{tabular}{|c|c|c|c|c|}
\hline
Noise Type & Noise Ratio & \begin{tabular}[c]{@{}c@{}}Ours w/o\\ MixUp Pred.\end{tabular} & \begin{tabular}[c]{@{}c@{}}Ours\\ (fix $\lambda=1$)\end{tabular} & Ours  \\ \hline
           & 0.3         & 84.37(-6.05)                                                   & 90.16(-0.26)                                                     & 90.42 \\
           & 0.4         & 78.33(-11.40)                                                  & 89.78(+0.05)                                                     & 89.73 \\
Symmetric  & 0.5         & 69.24(-19.43)                                                  & 88.69(+0.02)                                                     & 88.67 \\
           & 0.6         & 55.93(-31.07)                                                  & 86.97(-0.03)                                                     & 87.00 \\
           & 0.7         & 41.53(-41.19)                                                  & 81.82(-0.90)                                                     & 82.72 \\ \hline
           & 0.2         & 87.39(-3.45)                                                   & 90.82(-0.02)                                                     & 90.84 \\
Pairwise   & 0.3         & 79.53(-10.74)                                                  & 90.23(-0.04)                                                     & 90.27 \\
           & 0.4         & 64.88(-23.46)                                                  & 87.33(-1.01)                                                     & 88.34 \\ \hline
\end{tabular}
}
\end{table}

\textbf{Self distillation loss terms}
As shown in Table \ref{tab:div_test}, we evaluate various self distillation loss terms. 
In Table \ref{tab:div_test}, w/o means removing the self distillation loss term, JS Div (L1 loss, or L2 loss) means minimizing the discrepancy of $q_1$ and $q_2$ (mentioned in section \ref{actualmethod}) by using the JS divergence (L1, or L2 distance respectively). 
Besides, we can maximize the similarity of the two predictions measured by cosine similarity, and our method to optimize this term is the same as BYOL \cite{DBLP:conf/iccv/Banani021}. 
The results indicate that self distillation can effectively achieve performance improvement when the noise level is extremely high.
Meanwhile, when the noise level is relatively low, the performance is not very sensitive to self distillation loss $\gamma*Loss_{reg}$ in Eq.\ref{JSdiv}. Actually, we use JS divergence with IID data partitioning and L1 loss with non-IID data partitioning for evaluation, respectively. 
\begin{table}[htb]
\caption{Test accuracy (\%) for various self distillation terms. }
\label{tab:div_test}
\resizebox{\linewidth}{!}{
\begin{tabular}{|c|c|ccccc|}
\hline
\multirow{2}{*}{Noise Type} & \multirow{2}{*}{Noise Ratio} & \multicolumn{5}{c|}{\begin{tabular}[c]{@{}c@{}}Self Distillation\\ Loss Term\end{tabular}}                                                                                                                                                                                                                         \\ \cline{3-7} 
                            &                              & \multicolumn{1}{c|}{w/o}   & \multicolumn{1}{c|}{\begin{tabular}[c]{@{}c@{}}JS\\ Div\end{tabular}} & \multicolumn{1}{c|}{\begin{tabular}[c]{@{}c@{}}L1\\ Loss\end{tabular}} & \multicolumn{1}{c|}{\begin{tabular}[c]{@{}c@{}}L2\\ Loss\end{tabular}} & \begin{tabular}[c]{@{}c@{}}Cosine\\ Similarity\end{tabular} \\ \hline
                            & 0.3                          & \multicolumn{1}{c|}{89.86} & \multicolumn{1}{c|}{\textbf{90.42}}                                   & \multicolumn{1}{c|}{89.86}                                             & \multicolumn{1}{c|}{89.90}                                             & 89.47                                                       \\
                            & 0.4                          & \multicolumn{1}{c|}{89.19} & \multicolumn{1}{c|}{\textbf{89.73}}                                   & \multicolumn{1}{c|}{89.62}                                             & \multicolumn{1}{c|}{89.51}                                             & 88.83                                                       \\
Symmetric                   & 0.5                          & \multicolumn{1}{c|}{88.31} & \multicolumn{1}{c|}{\textbf{88.67}}                                            & \multicolumn{1}{c|}{88.49}                                    & \multicolumn{1}{c|}{88.31}                                             & 87.89                                                       \\
                            & 0.6                          & \multicolumn{1}{c|}{86.83} & \multicolumn{1}{c|}{87.00}                                            & \multicolumn{1}{c|}{\textbf{87.07}}                                    & \multicolumn{1}{c|}{86.72}                                             & 85.33                                                       \\
                            & 0.7                          & \multicolumn{1}{c|}{78.70} & \multicolumn{1}{c|}{82.72}                                            & \multicolumn{1}{c|}{\textbf{83.22}}                                    & \multicolumn{1}{c|}{80.61}                                             & 80.87                                                       \\ \hline
                            & 0.2                          & \multicolumn{1}{c|}{90.51} & \multicolumn{1}{c|}{\textbf{90.84}}                                            & \multicolumn{1}{c|}{90.37}                                             & \multicolumn{1}{c|}{90.56}                                    & 88.59                                                       \\
Pairwise                    & 0.3                          & \multicolumn{1}{c|}{89.61} & \multicolumn{1}{c|}{\textbf{90.27}}                                            & \multicolumn{1}{c|}{89.71}                                             & \multicolumn{1}{c|}{89.86}                                    & 86.42                                                       \\
                            & 0.4                          & \multicolumn{1}{c|}{83.10} & \multicolumn{1}{c|}{\textbf{88.34}}                                   & \multicolumn{1}{c|}{88.00}                                             & \multicolumn{1}{c|}{85.48}                                             & 82.31                                                       \\ \hline
\end{tabular}
}
\end{table}

\subsection{Experiments on Non-IID Data}
For completeness, we also evaluate the effectiveness of our method on non-IID data with extreme noisy labels on Fashion-MNIST dataset. 
As shown in Table \ref{tab:noniid}, our method outperforms other baselines on both IID and non-IID data with extreme noisy labels, but is worse than Co-teaching in the non-IID case. 
However, Co-teaching is hard to be applied in practice, since it needs to priorly determine a sensitive hyperparameter $R(T)$.
\begin{table}[htb]
\caption{Test accuracy (\%) for learning on both the IID and non-IID Fashion-MNIST dataset with extreme noisy labels.}
\label{tab:noniid}
\resizebox{\linewidth}{!}{
\begin{tabular}{|c|c|c|c|}
\hline
Method &
  Data Partitioning &
  \begin{tabular}[c]{@{}c@{}}Symmetric \\ ($\epsilon$=0.7)\end{tabular} &
  \begin{tabular}[c]{@{}c@{}}Pairwise\\ ($\epsilon$=0.4)\end{tabular} \\ \hline
\multirow{2}{*}{FedAvg}                    & IID     & 35.87          & 63.05          \\
                                           & Non-IID & 32.23          & 57.03          \\ \hline
\multirow{2}{*}{Symmetric CE}              & IID     & 49.34          & 68.44          \\
                                           & Non-IID & 45.59          & 66.21          \\ \hline
\multirow{2}{*}{Co-teaching}               & IID     & 81.03          & 83.03          \\
                                           & Non-IID & \textbf{81.32} & \textbf{82.29} \\ \hline
\multirow{2}{*}{Robust Federated Learning} & IID     & 79.22          & 88.17          \\
                                           & Non-IID & 74.89          & 70.45          \\ \hline
\multirow{2}{*}{FedAvg + LSR} & IID     & \textbf{82.72} & \textbf{88.34} \\
                                           & Non-IID & 80.64          & 76.30          \\ \hline
\end{tabular}
}
\end{table}

\subsection{Potential to Generalize}
We then verify the potential to suitably generalize our method LSR to other methods. 
For Co-teaching, each network in Co-teaching selects the samples into a correctly-labeled set according to CE loss computed by Eq.\ref{CE}, and feeds the correctly-labeled set to its peer network.
We simply modify the output logits of each peer network into a sharpened prediction version to compute CE loss to further guide sample-level selection and loss modeling, and other details are the same with the original Co-teaching. 
As shown in Table \ref{tab:general}, we can notably improve the performance of Co-teaching against extreme noisy labels by suitably introducing our method which implicitly regularize the local training process by enhancing the model prediction confidence. 
For Symmetric CE, by conducting data augmentation, we obtain original sample $x$ and augmented $\hat{x}$. 
By feeding them into network, we obtain the corresponding output logits $o_1$ and $o_2$. 
Obeying the original method of Symmetric CE, we just mix these logits up to compute Symmetric CE loss. 
The self distillation loss term is also added to explicitly regularize the local training process. 
The results indicate that Symmetric CE combined with our method can also achieve performance improvement. 
\label{realworldtest}
\begin{table}[htb]
\caption{Test accuracy (\%) of existing works combined with our method on Fashion-MNIST dataset.}
\label{tab:general}
\begin{tabular}{|c|c|c|}
\hline
Method                  & \begin{tabular}[c]{@{}c@{}}Symmetric\\  ($\epsilon$=0.7)\end{tabular} & \begin{tabular}[c]{@{}c@{}}Pairwise\\ ($\epsilon$=0.4)\end{tabular} \\ \hline
Co-teaching             & 81.03                                                                           & 83.03                                                                          \\ \hline
Co-teaching + LSR  & \textbf{85.76 (+ 4.73)}                                                         & \textbf{89.35 (+ 6.32)}                                                        \\ \hline
Symmetric CE            & 49.34                                                                           & 68.44                                                                          \\ \hline
Symmetric CE + LSR & \textbf{72.81 (+ 23.47)}                                                        & \textbf{77.33 (+ 8.89)}                                                        \\ \hline
\end{tabular} 
\end{table}

\subsection{Performance on the Real-world Dataset}
For evaluation on Clothing1M dataset, we exploit mixed precision technique \cite{micikevicius2018mixed} to accelerate training on this large dataset. 
For preprocessing, all images are resized to $224\times224$ and normalized. 
We selects two clients to train in each round (40 rounds in total). 
The results are shown in Table \ref{tab:testonclothing}. 
For \#5, \#7, \#8, the learning rate is set to 0.01. 
For \#6, the learning rate is set to 0.1 for the first 20 rounds, and 0.01 for the last 20 rounds. 
For \#6 and \#8, $\gamma$ is set to 1.2 and 1, respectively. 
For Robust FL, $T_{pl}$ is set to 10, and other hyperparameter setting follows \cite{yang2022robust}. 
We don't conduct experiment with Co-teaching, since its hyperparameter is hard to estimate in the practical (discussed in section \ref{results}). 
We also compares with \#1, \#2, \#3, \#4, which also focus on optimizing the training process in data centralized setting. 
The experimental results verify that our method combined with Symmetric CE can effectively improve the performance on Clothing1M dataset in federated setting.
\begin{table}[htb]
\caption{Average (3 trials) of best test accuracy on Clothing1M with non-IID partitioning. \#1,\#2 are quoted from \cite{DBLP:conf/cvpr/PatriniRMNQ17}, and \#3 and \#4 are quoted from \cite{DBLP:conf/nips/ZhangS18} and \cite{DBLP:conf/iccv/0001MCLY019}, respectively. C.L. and F.L. represent the data centralized learning and federated learning.}
\label{tab:testonclothing}
\resizebox{\linewidth}{!}{
\begin{tabular}{c|c|c|c}
\hline
\textbf{\#} & \textbf{Method}                 & \textbf{Setting} & \textbf{Test Accuracy(\%)} \\ \hline
1           & Cross Entropy                   & C. L.            & 68.94                           \\ \hline
2           & Symmetric CE                  & C. L.            & 71.02                          \\ \hline
3           & Forward                         & C. L.            & 69.84                           \\ \hline
4           & Generalized Cross Entropy       & C. L.            & 69.75                           \\ \hline
5           & FedAvg                          & F. L.            & 68.56                           \\ \hline
6           & FedAvg + LSR       & F. L.            & 69.30                           \\ \hline
7           & Symmetric CE                    & F. L.            & 69.63                           \\ \hline
8           & Symmetric CE + LSR & F. L.            & 70.46                           \\ \hline
9           & Robust Federated Learning       & F. L.            & 70.32               \\ \hline
\end{tabular}
}
\end{table}

\section{Discussion}
\begin{figure}[htb]
    \centering
    \includegraphics[width=\linewidth]{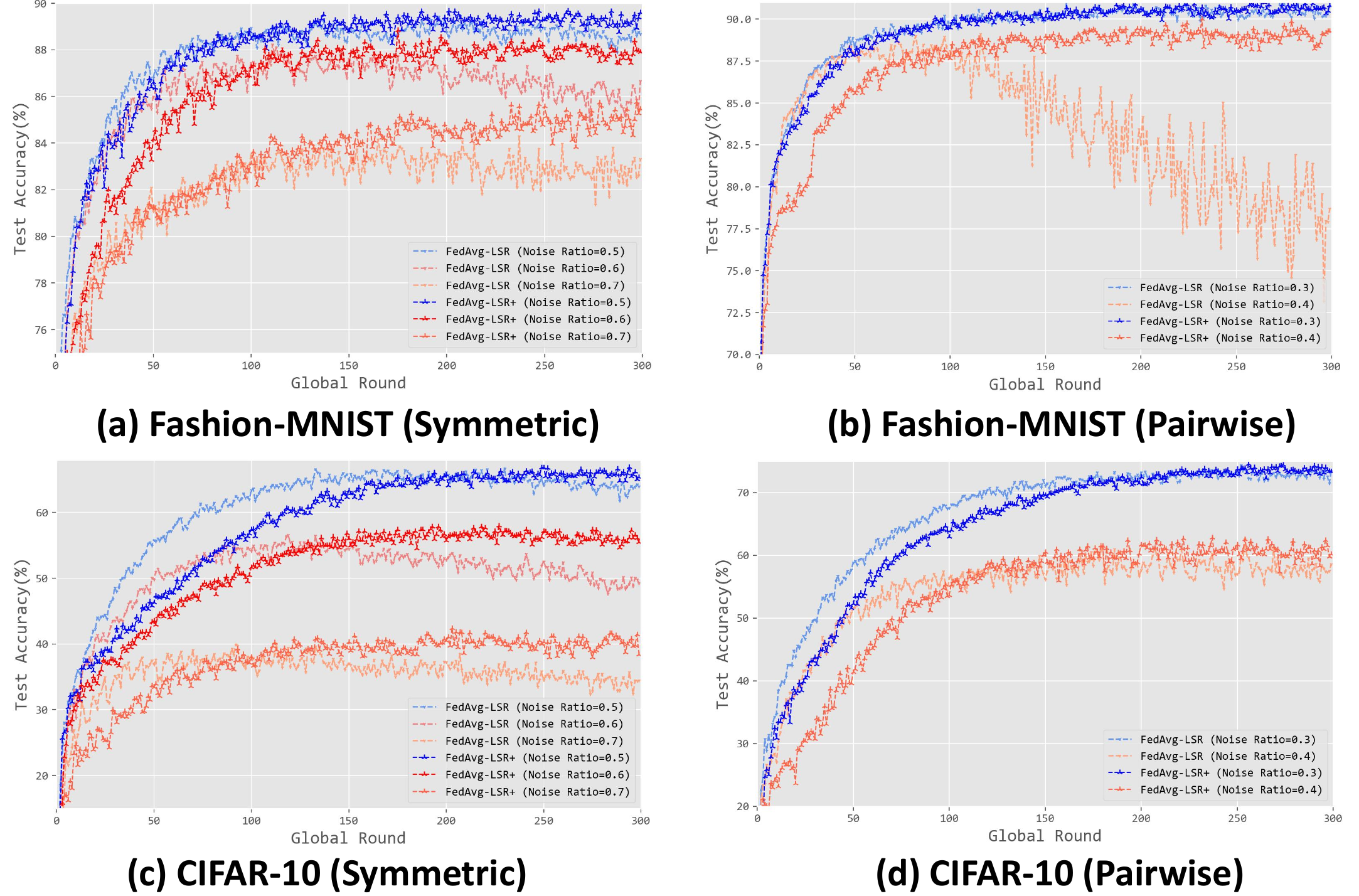}
    \caption{Test accuracy curve for the original and modified version of the proposed method.}
    \label{fig:extend}
\end{figure}
To further observe the stability and robustness for the proposed method FedAvg+LSR, we extend the global training round to 300. As shown in Figure \ref{fig:extend}, the original FedAvg+LSR is robust when the noise level is relatively lower, but shows degradation in extreme cases. Thus, we further adjust the orginal method (i.e. LSR) to a modified version (i.e. LSR+) by conducting entropy regularization on $o_1$ and $o_2$ (discussed in section \ref{actualmethod}), which aims to further strengthen the model discrimination confidence. The modified local training loss function can be formulated as:
\begin{equation}
    Loss_{LSR+} = Loss_{cls} + \gamma * Loss_{reg} + \lambda_{e} * L_{e}.
\end{equation}

For derivative works which adopt this method as a baseline, considering various noise levels, it is recommended that the trade-off coefficients $\gamma$ and $\lambda_e$ is mildly fixed to 0.4 and 0.6. The warm-up $t_w$ is set to $20\%$ of total training rounds. For self distillation, L1 loss or JS loss is preferred with fixed $T_d=\frac{1}{3}$.

\section{Conclusions}
\label{conclude}

In this work, we focus on FL against noisy-labeled data, and devote to alleviating the negative effect caused by noisy labels, which is an underlying challenge when deploying a practical FL system (e.g. a federated medical analysis system). 
We propose a Local Self-Regularization method to implicitly hinder the trained model from overfitting the noisy labels, and further leverage self knowledge distillation technique to explicitly regularize the model output discrepancy between original and mildly augmented instances.
Extensive experiments demonstrate that our proposed method is resistant to various noise levels on benchmark datasets. In addition, we integrate our proposed method with existing state-of-the-arts to achieve superior performance on real-world Clothing1M dataset. 
We believe there are many future works left to explore and address more practical issues hidden within a real FL system.
\begin{acks}
We sincerely thank reviewers for their careful reading, insightful comments and suggestions that lead to improve the quality of this manuscript. This work is supported by the National Natural Science Foundation of China (No. 62072436, No. 61732017, and No. 61872028).
\end{acks}

\bibliographystyle{ACM-Reference-Format}
\bibliography{sample-base}







\end{document}